\documentclass{article}

\usepackage[final]{nips_2016}

\usepackage[utf8]{inputenc} 
\usepackage[T1]{fontenc}    

\usepackage{url}            
\usepackage{color}
\usepackage{soul}
\usepackage{booktabs}       
\usepackage{amsfonts}       
\usepackage{nicefrac}       
\usepackage{microtype}      
\usepackage{amsmath}
\usepackage{graphicx} 
\usepackage{algorithmicx}
\usepackage{algorithm}

\usepackage{latexsym}
\usepackage{amsfonts}
\usepackage{algpseudocode}
\usepackage{array}
\usepackage{relsize}
\usepackage{soul}

\usepackage{helvet}
\usepackage{courier}
\usepackage{amsmath}
\usepackage{enumitem}

\usepackage{subfigure}

\newcommand{\vect}[1]{\mathbf{#1}}

\newcommand*{\affaddr}[1]{#1} 
\newcommand*{\affmark}[1][*]{\textsuperscript{#1}}
\newcommand*{\email}[1]{\texttt{#1}}

\begin{document}
%
\title{Exploring Question Understanding and Adaptation in Neural-Network-Based Question Answering}

\author{%
Junbei Zhang\affmark[1], Xiaodan Zhu\affmark[2], Qian Chen\affmark[1], Lirong Dai\affmark[1], Si Wei\affmark[3] and Hui Jiang\affmark[4]\\
\affaddr{\affmark[1]University of Science and Technology of China}\\
\affaddr{\affmark[2]National Research Council Canada}\\
\affaddr{\affmark[3]iFLYTEK Research, China}\\
\affaddr{\affmark[4]York University}\\
\email{\{zjunbei,cq1231\}@mail.ustc.edu.cn, xiaodan.zhu@nrc-cnrc.gc.ca}\\
\email{lrdai@ustc.edu.cn, siwei@iflytek.com, hj@cse.yorku.ca}\\
}

\maketitle
\begin{abstract}

The last several years have seen intensive interest in exploring neural-network-based models for machine comprehension (MC) and question answering (QA). In this paper, we approach the problems by closely modelling questions in a neural network framework. We first introduce syntactic information to help encode questions. We then view and model different types of questions and the information shared among them as an adaptation task and proposed adaptation models for them. On the Stanford Question Answering Dataset (SQuAD), we show that these approaches can help attain better results over a competitive baseline. 
\end{abstract}

\noindent \section{Introduction}

Enabling computers to understand given documents and answer questions about their content has recently attracted intensive interest, including but not limited to the efforts as in~\citep{richardson2013mctest,hermann2015teaching,hill2015goldilocks,rajpurkar2016squad,Nguyen2016MS,Berant2014Modeling}. Many specific problems such as machine comprehension and question answering often involve modeling such question-document pairs.

The recent availability of relatively large training datasets (see Section~\ref{sec:related} for more details) has made it more feasible to train and estimate rather complex models in an end-to-end fashion for these problems, in which a whole model is fit directly with given question-answer tuples and the resulting model has shown to be rather effective.  

In this paper, we take a closer look at modeling questions in such an end-to-end neural network framework, since we regard question understanding is of importance for such problems. We first introduced syntactic information to help encode questions. We then viewed and modelled different types of questions and the information shared among them as an adaptation problem and proposed adaptation models for them. On the Stanford Question Answering Dataset (SQuAD), we show that these approaches can help attain better results on our competitive baselines.

\section{Related Work}
\label{sec:related}
Recent advance on reading comprehension and question answering has been closely associated with the availability of various datasets. ~\cite{richardson2013mctest} released the MCTest data consisting of 500 short, fictional open-domain stories and 2000 questions. The CNN/Daily Mail dataset ~\citep{hermann2015teaching} contains news articles for close style machine comprehension, in which only entities are removed and tested for comprehension. Children's Book Test (CBT) ~\citep{hill2015goldilocks} leverages named entities, common nouns, verbs, and prepositions to test reading comprehension. The Stanford Question Answering Dataset (SQuAD) ~\citep{rajpurkar2016squad} is more recently released dataset, which consists of more than 100,000 questions for documents taken from Wikipedia across a wide range of topics. The question-answer pairs are annotated through crowdsourcing. Answers are spans of text marked in the original documents. In this paper, we use SQuAD to evaluate our models.

Many neural network models have been studied on the SQuAD task. ~\cite{wang2016machine} proposed ~\textit{match LSTM} to associate documents and questions and adapted the so-called~\textit{pointer Network}~\citep{vinyals2015pointer} to determine the positions of the answer text spans. ~\cite{yu2016end} proposed a~\textit{dynamic chunk reader} to extract and rank a set of answer candidates. ~\cite{yang2016words} focused on word representation and presented a fine-grained gating mechanism to dynamically combine word-level and character-level representations based on the properties of words. ~\cite{wang2016multi} proposed a~\textit{multi-perspective context matching} (MPCM) model, which matched an encoded document and question from multiple perspectives. ~\cite{xiong2016dynamic} proposed a dynamic decoder and so-called ~\textit{highway maxout network} to improve the effectiveness of the decoder. The~\textit{bi-directional attention flow} (BIDAF) ~\citep{seo2016bidirectional} used the bi-directional attention to obtain a question-aware context representation.

In this paper, we introduce syntactic information to encode questions with a specific form of recursive neural networks~\citep{zhu2015long,Tai2015,chen2016enhancing,Socher2011Parsing}. More specifically, we explore a tree-structured LSTM ~\citep{zhu2015long,Tai2015} which extends the linear-chain long short-term memory (LSTM)~\citep{hochreiter1997long} to a recursive structure, which has the potential to capture long-distance interactions over the structures.

Different types of questions are often used to seek for different types of information. For example, a "what" question could have very different property from that of a "why" question, while they may share information and need to be trained together instead of separately. We view this as a "adaptation" problem to let different types of questions share a basic model but still discriminate them when needed. Specifically, we are motivated by the ideas "i-vector"~\citep{dehak2011front} in speech recognition, where neural network based adaptation is performed among different (groups) of speakers and we focused instead on different types of questions here.

\section{The Approach}
\subsection{The Baseline Model}
Our baseline model is composed of the following typical components: \textit{word embedding}, \textit{input encoder}, \textit{alignment}, \textit{aggregation}, and \textit{prediction}. Below we discuss these components in more details.

\begin{figure}[h]
	\centering
	\includegraphics[width=0.85\linewidth]{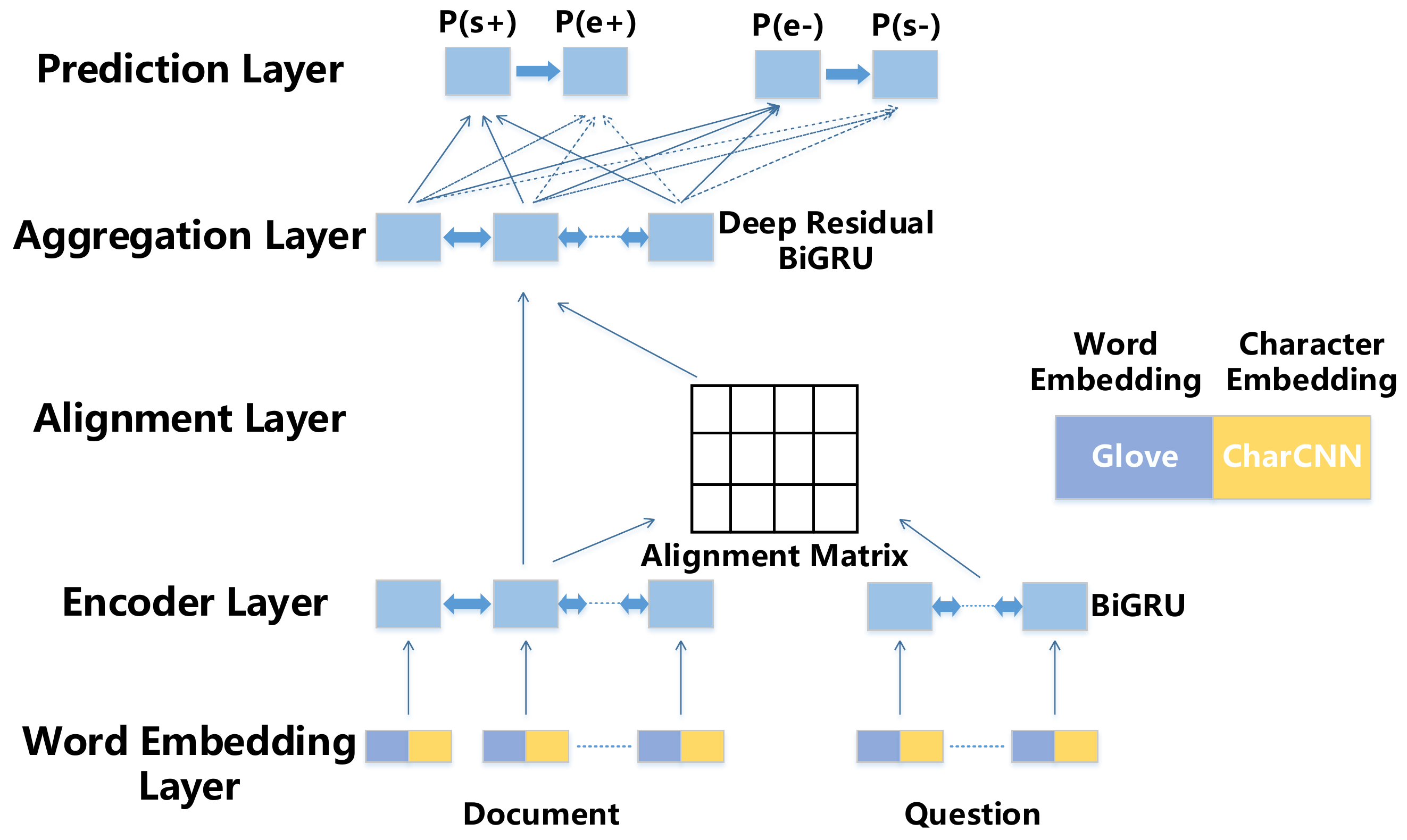}
	\caption{A high level view of our basic model.}
	\label{fig:squad}
\end{figure}

\paragraph{Word embedding}
We concatenate embedding at two levels to represent a word: the character composition and word-level embedding. The character composition feeds all characters of a word into a convolutional neural network (CNN)~\citep{kim2014convolutional} to obtain a representation for the word. And we use the pre-trained 300-D GloVe vectors~\citep{pennington2014glove} (see the experiment section for details) to initialize our word-level embedding. Each word is therefore represented as the concatenation of the character-composition vector and word-level embedding. This is performed on both questions and documents, resulting in two matrices: the $\vect{Q}^e \in \mathbb{R} ^{N\times d_w}$ for a question and the $\vect{D}^e \in \mathbb{R} ^{M\times d_w}$ for a document, where $N$ is the question length (number of word tokens), $M$ is the document length, and $d_w$ is the embedding dimensionality.

\paragraph{Input encoding}
The above word representation focuses on representing individual words, and an input encoder here employs recurrent neural networks to obtain the representation of a word under its context. We use bi-directional GRU (BiGRU)~\citep{cho2014properties} for both  documents and questions. 

\vspace{-2mm}
{\fontsize{10pt}{1.0cm}
 	\begin{align}
         \label{equ:blstm:a}
 		{\vect{Q}^c_i}&=\text{BiGRU}(\vect{Q}^e_i,i),\forall i \in [1, \dots, N] \\
        {\vect{D}^c_j}&=\text{BiGRU}(\vect{D}^e_j,j),\forall j \in [1, \dots, M]
         \label{equ:blstm:b}
 	\end{align}
}
\vspace{-6mm}

A BiGRU runs a forward and backward GRU on a sequence starting from the left and the right end, respectively. By concatenating the hidden states of these two GRUs for each word, we obtain the a representation for a question or document: $\vect{Q}^c \in \mathbb{R} ^{N\times d_c}$ for a question and $\vect{D}^c \in \mathbb{R} ^{M\times d_c}$ for a document.

\paragraph{Alignment}
Questions and documents interact closely. As in most previous work, our framework use both soft attention over questions and that over documents to capture the interaction between them. More specifically, in this soft-alignment layer, we first feed the contextual representation matrix $\vect{Q}^c$ and $\vect{D}^c$ to obtain alignment matrix $\vect{U} \in \mathbb{R} ^{N\times M}$:
\begin{equation}
\vect{U}_{ij} =\vect{Q}_i^c \cdot \vect{D}_j^{c\mathrm{T}}, \forall i \in [1, \dots, N], \forall j \in [1, \dots, M]
\end{equation}
Each $\vect{U}_{ij}$ represents the similarity between a question word $\vect{Q}_i^c$ and a document word $\vect{D}_j^c$.

\textit{\underline{\smash{Word-level Q-code}}}
Similar as in~\citep{seo2016bidirectional}, we obtain a word-level Q-code. Specifically, for each document word $w_j$, we find which words in the question are relevant to it. To this end, $\vect{a}_j\in \mathbb{R} ^{N}$ is computed with the following equation and used as a soft attention weight:
\begin{equation}
\vect{a}_j = softmax(\vect{U}_{:j}), \forall j \in [1, \dots, M]
\end{equation}
With the attention weights computed, we obtain the encoding of the question for each document word $w_j$ as follows, which we call~\textit{word-level Q-code} in this paper:
\begin{equation}
\vect{Q}^w=\vect{a}^{\mathrm{T}} \cdot \vect{Q}^{c} \in \mathbb{R} ^{M\times d_c}
\end{equation}

\textit{\underline{\smash{Question-based filtering}}} To better explore question understanding, we design this~\textit{question-based filtering} layer. As detailed later, different question representation can be easily incorporated to this layer in addition to being used as a filter to find key information in the document based on the question. This layer is  expandable with more complicated question modeling. 

In the basic form of question-based filtering, for each question word $w_i$, we find which words in the document are associated. Similar to $\vect{a}_j$ discussed above, we can obtain the attention weights on document words for each question word $w_i$:
\begin{equation}
\vect{b}_i=softmax(\vect{U}_{i:})\in \mathbb{R} ^{M}, \forall i \in [1, \dots, N]
\end{equation}
By pooling $\vect{b}\in \mathbb{R} ^{N\times M}$, we can obtain a question-based filtering weight $\vect{b}^f$:
\begin{equation}
\vect{b}^f=norm(pooling(\vect{b})) \in \mathbb{R} ^{M}
\end{equation}
\begin{equation}
norm(\vect{x})=\frac{\vect{x}}{\sum_i x_i}
\end{equation}
where the specific pooling function we used include max-pooling and mean-pooling. Then the document softly filtered based on the corresponding question  $\vect{D}^f$ can be calculated by:
\begin{equation}
\vect{D}_j^{f_{max}}=b^{f_{max}}_j \vect{D}_j^{c},  \forall j \in [1, \dots, M]
\end{equation}
\begin{equation}
\vect{D}_j^{f_{mean}}=b^{f_{mean}}_j \vect{D}_j^{c},  \forall j \in [1, \dots, M]
\end{equation}
\begin{equation}
\vect{D}^f= [\vect{D}^{f_{max}}, \vect{D}^{f_{mean}}]
\end{equation}
Through concatenating the document representation $\vect{D}^c$, word-level Q-code $\vect{Q}^w$ and question-filtered document $\vect{D}^f$, we can finally obtain the alignment layer representation:
\begin{equation}
\vect{I}=[\vect{D}^c, \vect{Q}^w,\vect{D}^c \circ \vect{Q}^w,\vect{D}^c - \vect{Q}^w, \vect{D}^f, \vect{b}^{f_{max}}, \vect{b}^{f_{mean}}] \in \mathbb{R} ^{M \times (6d_c+2)}
\end{equation}
where "$\circ$" stands for element-wise multiplication and "$-$" is simply the vector subtraction.

\begin{figure}[h]
	\centering
	\includegraphics[width=0.4\linewidth]{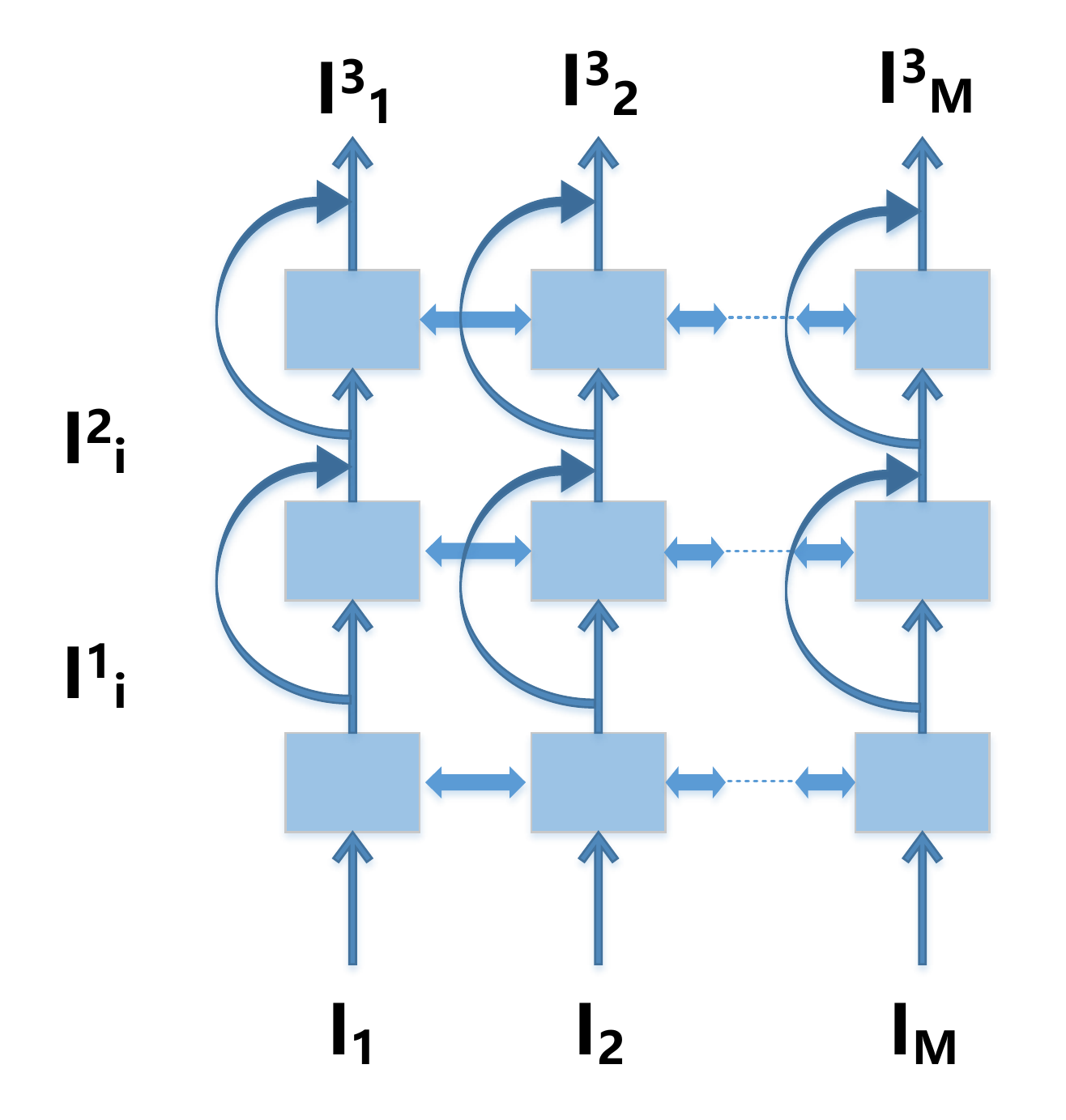}
	\caption{The inference layer implemented with a residual network.}
	\label{fig:inference}
\end{figure}

\paragraph{Aggregation}
After acquiring the local alignment representation, key information in document and question has been collected, and the aggregation layer is then performed to find answers. We use three BiGRU layers to model the process that aggregates local information to make the global decision to find the answer spans. We found a residual architecture~\citep{he2016deep} as described in Figure~\ref{fig:inference} is very effective in this aggregation process:
\begin{equation}
\vect{I}^1_i=\text{BiGRU}(\vect{I}_i)
\end{equation}
\begin{equation}
\vect{I}^2_i=\vect{I}^1_i + \text{BiGRU}(\vect{I}^1_i)
\end{equation}
\begin{equation}
\vect{I}^3_i=\vect{I}^2_i + \text{BiGRU}(\vect{I}^2_i)
\end{equation}

\paragraph{Prediction}
The SQuAD QA task requires a span of text to answer a question. We use a pointer network~\citep{vinyals2015pointer} to predict the starting and end position of answers as in~\citep{wang2016machine}. Different from their methods, we use a two-directional prediction to obtain the positions. For one direction, we first predict the starting position of the answer span followed by predicting the end position, which is implemented with the following equations:
\begin{equation}
P(s+)=softmax(W_{s+}\cdot I^3)
\end{equation}
\begin{equation}
P(e+)=softmax(W_{e+} \cdot I^3  + W_{h+} \cdot h_{s+})
\end{equation}
where $\vect{I}^3$ is inference layer output, $\vect{h}_{s+}$ is the hidden state of the first step, and all $\vect{W}$ are trainable matrices. We also perform this by predicting the end position first and then the starting position:
\begin{equation}
P(e-)=softmax(W_{e-}\cdot I^3)
\end{equation}
\begin{equation}
P(s-)=softmax(W_{s-} \cdot I^3  + W_{h-} \cdot h_{e-})
\end{equation}
We finally identify the span of an answer with the following equation:
\begin{equation}
P(s)=pooling([P(s+), P(s-)])
\end{equation}
\begin{equation}
P(e)=pooling([P(e+), P(e-)])
\end{equation}
We use the mean-pooling here as it is more effective on the development set than the alternatives such as the max-pooling.

\subsection{Question Understanding and Adaptation}

\subsubsection{Introducing syntactic information for neural question encoding}

The interplay of syntax and semantics of natural language questions is of interest for question representation. We attempt to incorporate syntactic information in questions representation with TreeLSTM~\citep{zhu2015long,Tai2015}. In general a TreeLSTM could perform semantic composition over given syntactic structures. 

Unlike the chain-structured LSTM~\citep{hochreiter1997long}, the TreeLSTM captures long-distance interaction on a tree. The update of a TreeLSTM node is described at a high level with Equation (\ref{equ:TreeLSTM}), and the detailed computation is described in (\ref{equ:begin}--\ref{equ:end}). Specifically, the input of a TreeLSTM node is used to configure four gates: the input gate $\vect{i}_t$, output gate $\vect{o}_t$, and the two forget gates $\vect{f}_t^L$ for the left child input and $\vect{f}_t^R$ for the right. The memory cell $\vect{c}_t$ considers each child's cell vector, $\vect{c}_{t-1}^L$ and $\vect{c}_{t-1}^R$, which are gated by the left forget gate $\vect{f}_t^L$ and right forget gate $\vect{f}_t^R$, respectively. 

\vspace{-4mm}
{\fontsize{10pt}{1.0cm}
	\begin{align}
    \label{equ:TreeLSTM}
        \vect{h}_t &= \text{TreeLSTM}(\vect{x}_t, \vect{h}_{t-1}^L, \vect{h}_{t-1}^R), \\
		\label{equ:begin}
		\vect{h}_t &= \vect{o}_t \circ \tanh(\vect{c}_{t}),\\
		\vect{o}_t  &= \sigma (\vect{W}_o \vect{x}_t + \vect{U}_o^L \vect{h}_{t-1}^L +  \vect{U}_o^R \vect{h}_{t-1}^R), \\
		\vect{c}_t &= \vect{f}_t^L \circ \vect{c}_{t-1}^L + \vect{f}_t^R \circ \vect{c}_{t-1}^R  + \vect{i}_t \circ \vect{u}_t, \\
		\vect{f}_t^L  &= \sigma (\vect{W}_f \vect{x}_t + \vect{U}_f^{LL} \vect{h}_{t-1}^L +  \vect{U}_f^{LR} \vect{h}_{t-1}^R),\\
\vect{f}_t^R  &= \sigma (\vect{W}_f \vect{x}_t + \vect{U}_f^{RL} \vect{h}_{t-1}^L +  \vect{U}_f^{RR} \vect{h}_{t-1}^R), \\
		\vect{i}_t  &= \sigma (\vect{W}_i \vect{x}_t + \vect{U}_i^L \vect{h}_{t-1}^L +  \vect{U}_i^R \vect{h}_{t-1}^R), \\
		\vect{u}_t  &= \tanh (\vect{W}_c \vect{x}_t + \vect{U}_c^L \vect{h}_{t-1}^L +  \vect{U}_c^R \vect{h}_{t-1}^R),
    \label{equ:end}
	\end{align}
}
\noindent where $\sigma$ is the sigmoid function, $\circ$ is the element-wise multiplication of two vectors, and all $\vect{W}$, $\vect{U}$ are trainable matrices.

To obtain the parse tree information, we use Stanford CoreNLP (PCFG Parser)~\citep{Manning2014The,Klein2003Accurate} to produce a binarized constituency parse for each question and build the TreeLSTM based on the parse tree. The root node of TreeLSTM is used as the representation for the whole question. More specifically, we use it as TreeLSTM Q-code $\vect{Q}^{TL}\in \mathbb{R} ^{d_c}$, by not only simply concatenating it to the alignment layer output but also using it as a question filter, just as we discussed in the \textit{\smash{question-based filtering}} section:
\begin{equation}
\vect{Q}^{TL}=\text{TreeLSTM}(\vect{Q}^e) \in \mathbb{R} ^{d_c}
\end{equation}
\begin{equation}
\vect{b}^{TL}=norm(\vect{Q}^{TL} \cdot \vect{D}^{c\mathrm{T}})  \in \mathbb{R} ^{M}
\end{equation}
\begin{equation}
\vect{D}_j^{TL}= b_j^{TL}  \vect{D}_j^{c},  \forall j \in [1, \dots, M]
\end{equation}
\begin{equation}
\vect{I}_{new}=[\vect{I}, repmat(\vect{Q}^{TL}), \vect{D}^{TL}, \vect{b}^{TL}]
\end{equation}
where $\vect{I}_{new}$ is the new output of alignment layer, and function $repmat$ copies $\vect{Q}^{TL}$ for M times to fit with $\vect{I}$.

\subsubsection{Question Adaptation}
Questions by nature are often composed to fulfill different types of information needs. For example, a "when" question seeks for different types of information (i.e., temporal information) than those for a "why" question. Different types of questions and the corresponding answers could potentially have different distributional regularity. 

\paragraph{Explicit question-type embedding}
The previous models are often trained for all questions without explicitly discriminating different question types; however, for a target question, both the common features shared by all questions and the specific features for a specific type of question are further considered in this paper, as they could potentially obey different distributions. 
In this paper we further explicitly model different types of questions in the end-to-end training. We start from a simple way to first analyze the word frequency of all questions, and obtain top-10 most frequent question types: \textit{what, how, who, when, which, where, why, be, whose,} and \textit{whom}, in which \textit{be} stands for the questions beginning with different forms of the word~\textit{be} such as~\textit{is, am}, and ~\textit{are}. We explicitly encode question-type information to be an 11-dimensional one-hot vector (the top-10 question types and "other" question type). Each question type is with a trainable embedding vector. We call this explicit question type code, $\vect{ET}\in \mathbb{R} ^{d_{ET}}$. Then the vector for each question type is tuned during training, and is added to the system with the following equation:
\begin{equation}
\vect{I}_{new}=[\vect{I}, repmat(\vect{ET})]
\end{equation}

\paragraph{Question adaptation}
As discussed, different types of questions and their answers may share common regularity and have separate property at the same time. We also view this as an adaptation problem in order to let different types of questions share a basic model but still discriminate them when needed. Specifically, we borrow ideas from speaker adaptation~\citep{dehak2011front} in speech recognition, where neural-network-based adaptation is performed among different groups of speakers.

\begin{figure}[h]
	\centering
	\includegraphics[width=0.4\linewidth]{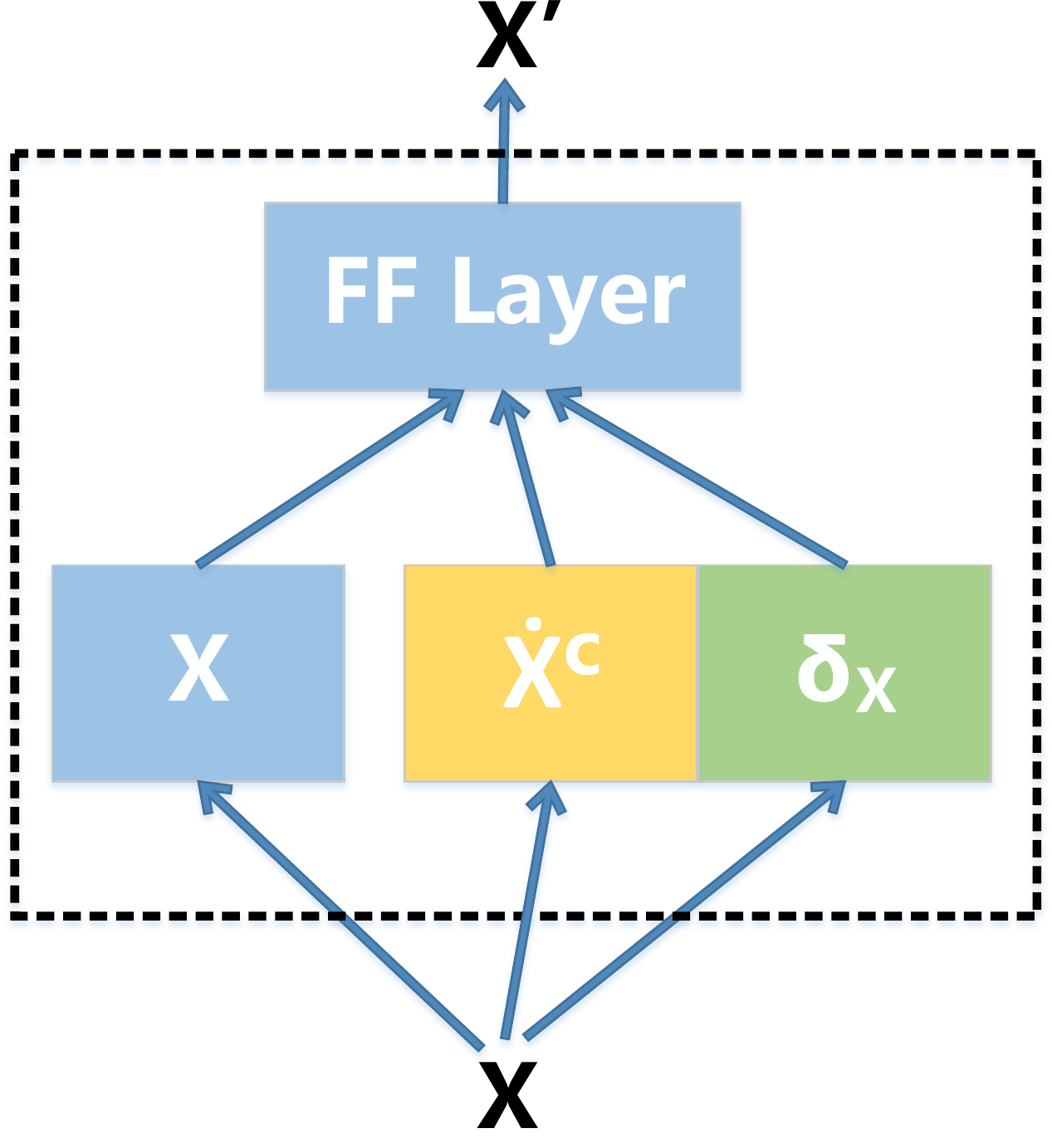}
	\caption{The~\textit{discriminative block} for question discrimination and adaptation.}
	\label{fig:discriminative_block}
\end{figure}

Conceptually we regard a type of questions as a group of acoustically similar speakers. Specifically we propose a question discriminative block or simply called a \textit{discriminative block} (Figure~\ref{fig:discriminative_block}) below to perform question adaptation. 
The main idea is described below:
\begin{equation}
\vect{x^\prime} = f([\vect{x}, \vect{\bar{x}}^c, \vect{\delta_x}])
\end{equation}

For each input question $\vect{x}$, we can decompose it to two parts: the cluster it belong(i.e., question type) and the diverse in the cluster. The information of the cluster is encoded in a vector $\vect{\bar{x}}^c$. In order to keep calculation differentiable, we compute the weight of all the clusters based on the distances of $\vect{x}$ and each cluster center vector, in stead of just choosing the closest cluster. Then the discriminative vector $\vect{\delta_x}$ with regard to these most relevant clusters are computed. All this information is combined to obtain the discriminative information. In order to keep the full information of input, we also copy the input question $\vect{x}$, together with the acquired discriminative information, to a feed-forward layer to obtain a new representation $\vect{x^\prime}$ for the question.

More specifically, the adaptation algorithm contains two steps: \textit{adapting} and \textit{updating}, which is detailed as follows:

\begin{itemize}
\setlength{\itemsep}{0pt}
\setlength{\parsep}{0pt}
\setlength{\parskip}{0pt}
\item \textbf{Adapting}
In the adapting step, we first compute the similarity score between an input question vector $\vect{x}\in \mathbb{R} ^{h}$ and each centroid vector of $K$ clusters$~\vect{\bar{x}}\in \mathbb{R} ^{K \times h}$. Each cluster here models a question type. Unlike the explicit question type modeling discussed above, here we do not specify what question types we are modeling but let the system to learn. Specifically, we only need to pre-specific how many clusters, $K$, we are modeling. The similarity between an input question and cluster centroid can be used to compute similarity weight $\vect{w}^a$:
\begin{equation}
w_k^a = softmax(cos\_sim(\vect{x}, \vect{\bar{x}}_k), \alpha),  \forall k \in [1, \dots, K]
\end{equation}
\begin{equation}
cos\_sim(\vect{u}, \vect{v}) = \frac{<\vect{u},\vect{v}>}{||\vect{u}|| \cdot ||\vect{v}||}
\end{equation}
\begin{equation}
softmax(x_i, \alpha) = \frac{e^{\alpha x_i}}{\sum_j e^{\alpha x_j}}
\end{equation}
We set $\alpha$ equals 50 to make sure only closest class will have a high weight while maintain differentiable. Then we acquire a soft class-center vector $\vect{\bar{x}}^c$:
\begin{equation}
\vect{\bar{x}}^c = \sum_k w^a_k \vect{\bar{x}}_k \in \mathbb{R} ^{h}
\end{equation}
We then compute a discriminative vector $\vect{\delta_x}$ between the input question with regard to the soft class-center vector:
\begin{equation}
\vect{\delta_x} = \vect{x} - \vect{\bar{x}}^c
\end{equation}

Note that $\bar{\vect{x}}^c$ here models the cluster information and $\vect{\delta_x}$ represents the discriminative information in the cluster. By feeding $\vect{x}$, $\bar{\vect{x}}^c$ and $\vect{\delta_x}$ into feedforward layer with Relu, we obtain $\vect{x'}\in \mathbb{R} ^{K}$:
\begin{equation}
\vect{x'} = Relu(\vect{W} \cdot [\vect{x},\bar{\vect{x}}^c,\vect{\delta_x}])
\end{equation}
With $\vect{x'}$ ready, we can apply Discriminative Block to any question code and obtain its adaptation Q-code. In this paper, we use TreeLSTM Q-code as the input vector $\vect{x}$, and obtain TreeLSTM adaptation Q-code $\vect{Q}^{TLa}\in \mathbb{R} ^{d_c}$. Similar to TreeLSTM Q-code $\vect{Q}^{TL}$, we concatenate $\vect{Q}^{TLa}$ to alignment output $\vect{I}$ and also use it as a question filter:
\begin{equation}
\vect{Q}^{TLa} = Relu(\vect{W} \cdot [\vect{Q}^{TL},\overline{\vect{Q}^{TL}}^c,\vect{\delta_{\vect{Q}^{TL}}}])
\end{equation}
\begin{equation}
\vect{b}^{TLa}=norm(\vect{Q}^{TLa} \cdot \vect{D}^{c\mathrm{T}})  \in \mathbb{R} ^{M}
\end{equation}
\begin{equation}
\vect{D}_j^{TLa}= b_j^{TLa}  \vect{D}_j^{c},  \forall j \in [1, \dots, M]
\end{equation}
\begin{equation}
\vect{I}_{new}=[\vect{I}, repmat(\vect{Q}^{TLa}), \vect{D}^{TLa}, \vect{b}^{TLa}]
\end{equation}

\item \textbf{Updating}  The updating stage attempts to modify the center vectors of the $K$ clusters in order to fit each cluster to model different types of questions. The updating is performed according to the following formula:
\begin{equation}
\vect{\bar{x}'}_k = (1-\beta \text{w}_k^a)\vect{\bar{x}}_k+\beta \text{w}_k^a\vect{x},  \forall k \in [1, \dots, K]
\end{equation}
In the equation, $\beta$ is an updating rate used to control the amount of each updating, and we set it to 0.01. When $\vect{x}$ is far away from $K$-th cluster center $\vect{\bar{x}}_k$, $\text{w}_k^a$ is close to be value 0 and the $k$-th cluster center $\vect{\bar{x}}_k$ tends not to be updated. If $\vect{x}$ is instead close to the $j$-th cluster center $\vect{\bar{x}}_j$, $\text{w}_k^a$ is close to the value 1 and the centroid of the $j$-th cluster $\vect{\bar{x}}_j$ will be updated more aggressively using $\vect{x}$.
\end{itemize}

\section{Experiment Results}
\subsection{Set-Up}
We test our models on Stanford Question Answering Dataset (SQuAD)~\citep{rajpurkar2016squad}. The SQuAD dataset consists of more than 100,000 questions annotated by crowdsourcing workers on a selected set of Wikipedia articles, and the answer to each question is a span of text in the  Wikipedia articles. Training data includes 87,599 instances and validation set has 10,570 instances. The test data is hidden and kept by the organizer. The evaluation of SQuAD is Exact Match (EM) and F1 score.

We use pre-trained \textit{300-D Glove 840B} vectors~\citep{pennington2014glove} to initialize our word embeddings. Out-of-vocabulary (OOV) words are initialized randomly with Gaussian samples. CharCNN filter length is 1,3,5, each is 50 dimensions. All vectors including word embedding are updated during training. The cluster number K in discriminative block is 100. The Adam method~\citep{Kingma2014AdamAM} is used for optimization. And the first momentum is set to be 0.9 and the second 0.999. The initial learning rate is 0.0004 and the batch size is 32. We will half learning rate when meet a bad iteration, and the patience is 7. Our early stop evaluation is the EM and F1 score of validation set. All hidden states of GRUs, and TreeLSTMs are 500 dimensions, while word-level embedding $d_w$ is 300 dimensions. We set max length of document to 500, and drop the question-document pairs beyond this on training set. Explicit question-type dimension $d_{ET}$ is 50. We apply dropout to the Encoder layer and aggregation layer with a dropout rate of 0.5.   

\begin{table*}[t]
	\renewcommand{\arraystretch}{1.3}
	\centering
	\begin{tabular}{llll}
		\toprule
		Model    &  EM  &  F1  \\
		\midrule
        Logistic Regression Baseline ~\citep{rajpurkar2016squad} &  40.4  & 51.0 \\
		\midrule
        Match-LSTM with Ans-Ptr (Sentence)~\citep{wang2016machine}  & 54.505 & 67.748 \\
        Match-LSTM with Ans-Ptr (Boundary)~\citep{wang2016machine}  & 60.474 & 70.695 \\
        Dynamic Chunk Reader~\citep{yu2016end} & 62.499 & 70.956 \\
        Fine-Grained Gating~\citep{yang2016words}  & 62.446 & 73.327 \\
        Match-LSTM with Bi-Ans-Ptr (Boundary)~\citep{wang2016machine}  & 64.744 & 73.743 \\
        Multi-Perspective Matching~\citep{wang2016multi} & 65.551 & 75.118 \\ 
        Dynamic Coattention Networks~\citep{xiong2016dynamic}  & 66.233 & 75.896 \\
        BiLSTM [German Research Center for Artificial Intelligence(unpublished)]  & 68.436 & 77.070 \\
        BiDAF~\citep{seo2016bidirectional} & 67.974 & 77.323 \\
        \textbf{jNet(Ours)} & \textbf{68.730} & \textbf{77.393} \\ 
		r-net [Microsoft Research Asia(unpublished)] &  70.062  & 78.782 \\
		\midrule
        Human Performance~\citep{rajpurkar2016squad} &  82.304  & 91.221 \\
		\bottomrule
	\end{tabular}
	\caption{The official leaderboard of single models on SQuAD test set as we submitted our systems (January 20, 2017).}
    \label{tab:result}
\end{table*}

\subsection{Results}

\paragraph{Overall results}
Table~\ref{tab:result} shows the official leaderboard on SQuAD test set when we submitted our system. Our model achieves a 68.73\% EM score and 77.39\% F1 score, which is ranked among the state of the art single models (without model ensembling).

\begin{table*}[ht]
	\renewcommand{\arraystretch}{1.3}
	\centering
	\begin{tabular}{llll}
		\toprule
		Model    &  EM  &  F1 \\
		\midrule
		Baseline &  68.00  & 77.36   \\
        +Explicit question types ($\vect{ET}$) & 68.16 & 77.58 \\
        +TreeLSTM ($\vect{Q}^{TL}$) & 68.29 & 77.67 \\
        +TreeLSTM adaptation ($\vect{Q}^{TLa}$, K=20) & 68.73 & 77.74 \\
        \textbf{+TreeLSTM adaptation ($\vect{Q}^{TLa}$, K=100)} & \textbf{69.10} & \textbf{78.38} \\
		\bottomrule
	\end{tabular}
	\caption{Performance of various Q-code on the development set.}
    \label{tab:ablation}
\end{table*}

Table~\ref{tab:ablation} shows the ablation performances of various Q-code on the development set. Note that since the testset is hidden from us, we can only perform such an analysis on the development set. Our baseline model using no Q-code achieved a 68.00\% and 77.36\% EM and F1 scores, respectively. When we added the explicit question type T-code into the baseline model, the performance was improved slightly to 68.16\%(EM) and 77.58\%(F1). We then used TreeLSTM introduce syntactic parses for question representation and understanding (replacing simple question type as question understanding Q-code), which consistently shows further improvement. We further incorporated the soft adaptation. When letting the number of hidden question types ($K$) to be 20, the performance improves to 68.73\%/77.74\% on EM and F1, respectively,  which corresponds to the results of our model reported in Table~\ref{tab:result}.  Furthermore, after submitted our result, we have experimented with a large value of $K$ and found that when $K=100$, we can achieve a better performance of 69.10\%/78.38\% on the development set.

\begin{figure}[h!]
 \centering
 \subfigure[Question Type Results]{
   \label{fig:subfig:a} 
   \includegraphics[width=0.5\linewidth]{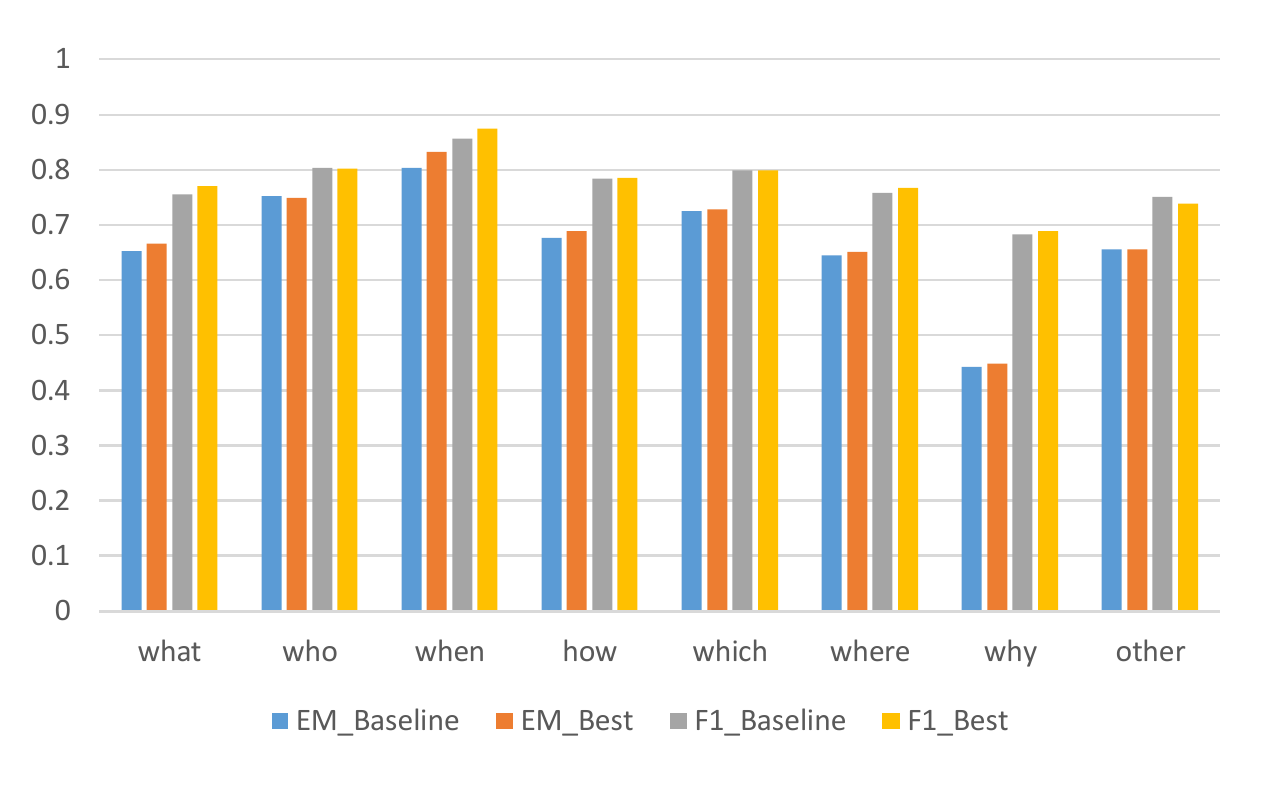}}
 \subfigure[Question Type Amount]{
   \label{fig:subfig:b} 
   \includegraphics[width=0.4\linewidth]{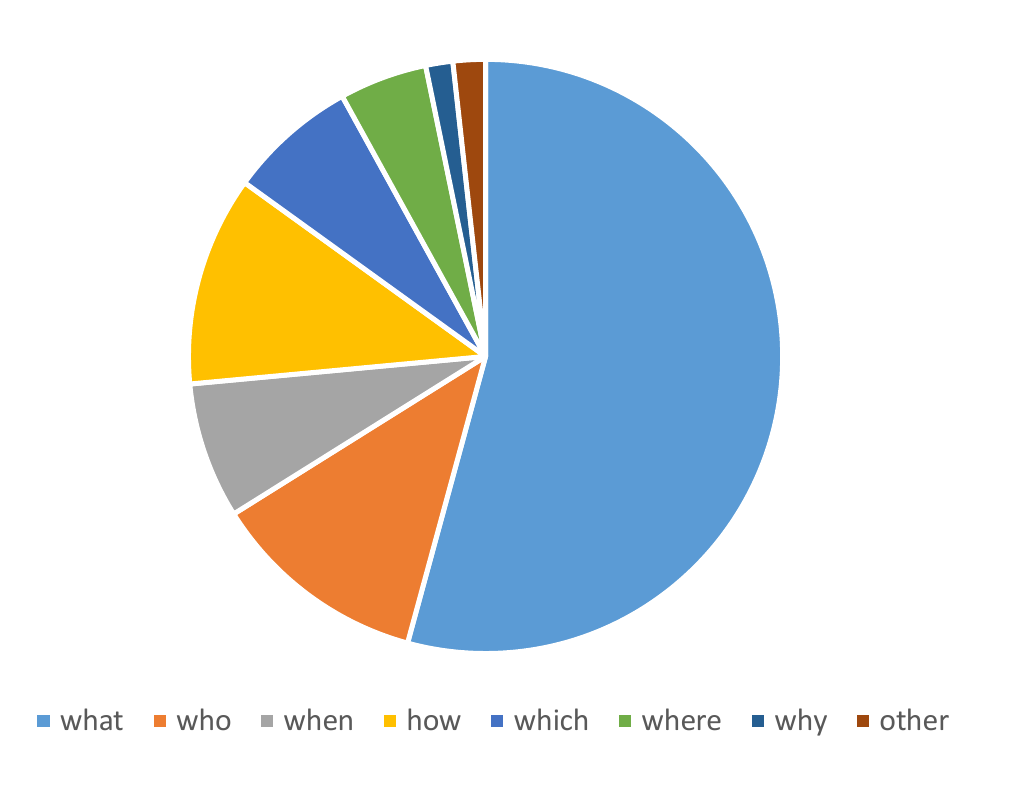}}
 \caption{Question Type Analysis}
 \label{fig:subfig} 
\end{figure}

Figure~\ref{fig:subfig:a} shows the EM/F1 scores of different question types while Figure~\ref{fig:subfig:b} is the question type amount distribution on the development set. In Figure~\ref{fig:subfig:a} we can see that the average EM/F1 of the "when" question is highest and those of the "why" question is the lowest. From Figure~\ref{fig:subfig:b} we can see the "what" question is the major class.

\begin{figure}[h!]
	\centering
	\includegraphics[width=0.6\linewidth]{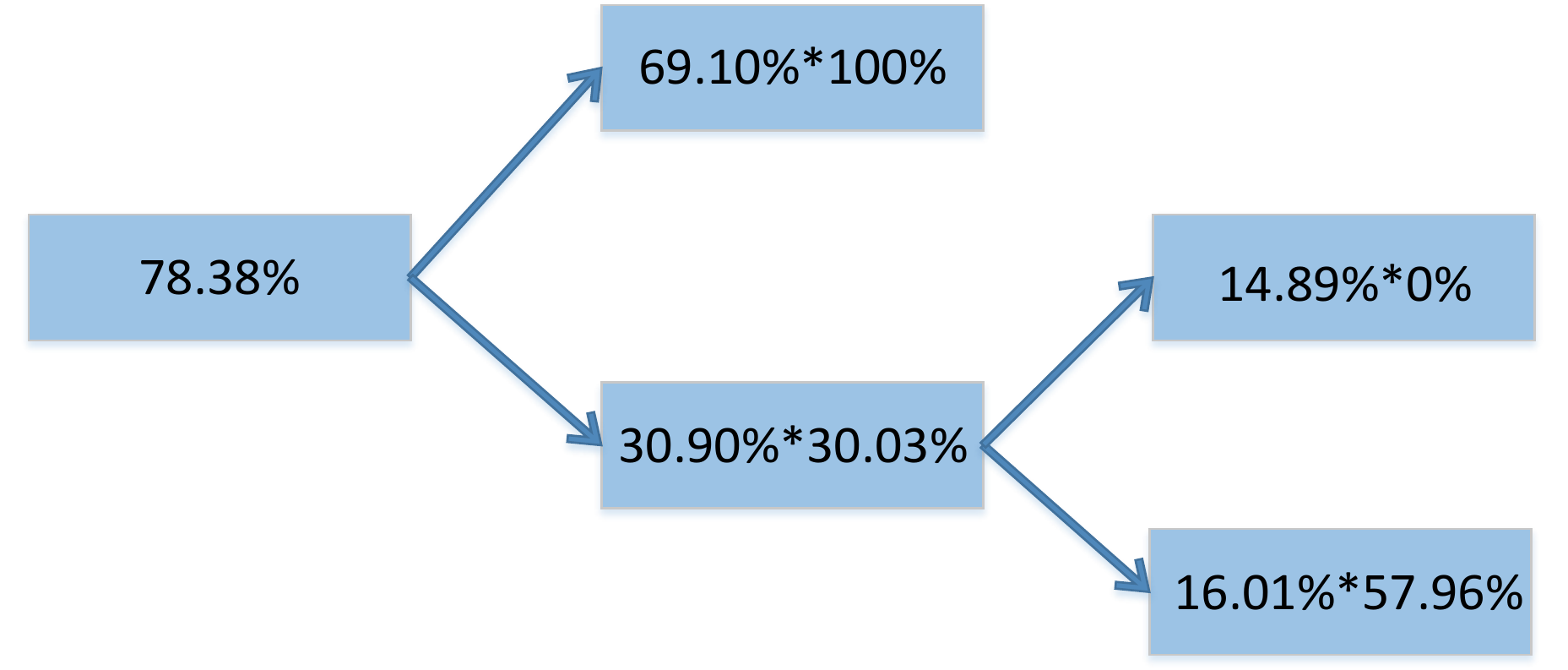}
	\caption{F1 Score Analysis.}
	\label{fig:result1}
\end{figure}

Figure~\ref{fig:result1} shows the composition of F1 score. Take our best model as an example, we observed a 78.38\% F1 score on the whole development set, which can be separated into two parts: one is where F1 score equals to 100\%, which means an exact match. This part accounts for 69.10\% of the entire development set. And the other part accounts for 30.90\%, of which the average F1 score is 30.03\%. For the latter, we can further divide it into two sub-parts: one is where the F1 score equals to 0\%, which means that predict answer is totally wrong. This part occupies 14.89\% of the total development set. The other part accounts for 16.01\% of the development set, of which average F1 score is 57.96\%. From this analysis we can see that reducing the zero F1 score (14.89\%) is potentially an important direction to further improve the system.

\section{Conclusions}
Closely modelling questions could be of importance for question answering and machine reading.  In this paper, we introduce syntactic information to help encode questions in neural networks. We view and model different types of questions and the information shared among them as an adaptation task and proposed adaptation models for them. On the Stanford Question Answering Dataset (SQuAD), we show that these approaches can help attain better results over a competitive baseline.

\bibliographystyle{plainnat}
\bibliography{reference}

\end{document}